\documentclass[conference]{IEEEtran}
\usepackage{graphicx}

\ifCLASSINFOpdf
\else
\fi
\begin{document}
%
\title{BioIE: Biomedical Information Extraction with Multi-head Attention Enhanced Graph Convolutional Network}

\author{\IEEEauthorblockN{Jialun Wu$^{1}$,
Yang Liu$^{1}$,
Zeyu Gao$^{1}$, 
Tieliang Gong$^{1}$,
Chunbao Wang$^{2}$,and
Chen Li$^{1,*}$}

\IEEEauthorblockA{$^{1}$School of Computer Science and Technology, Xi'an Jiaotong University, Xi'an, China\\
National Engineering Lab for Big Data Analytics, Xi'an Jiaotong University, Xi'an, China} 
\IEEEauthorblockA{$^{2}$Department of Pathology, the First Affiliated Hospital of Xi'an Jiaotong University, Xi'an, China\\
Email: andylun96@stu.xjtu.edu.cn}}

\maketitle

\begin{abstract}

Constructing large-scaled medical knowledge graphs (MKGs) can significantly boost healthcare applications for medical surveillance, bring much attention from recent research. An essential step in constructing large-scale MKG is extracting information from medical reports. Recently, information extraction techniques have been proposed and show promising performance in biomedical information extraction. However, these methods only consider limited types of entity and relation due to the noisy biomedical text data with complex entity correlations. Thus, they fail to provide enough information for constructing MKGs and restrict the downstream applications. To address this issue, we propose Biomedical Information Extraction (BioIE), a hybrid neural network to extract relations from biomedical text and unstructured medical reports. Our model utilizes a multi-head attention enhanced graph convolutional network (GCN) to capture the complex relations and context information while resisting the noise from the data. 
We evaluate our model on two major biomedical relationship extraction tasks, chemical-disease relation (CDR) and chemical–protein interaction (CPI), and a cross-hospital pan-cancer pathology report corpus. The results show that our method achieves superior performance than baselines. Furthermore, we evaluate the applicability of our method under a transfer learning setting and show that BioIE achieves promising performance in processing medical text from different formats and writing styles.

\end{abstract}


%
\IEEEpeerreviewmaketitle

\section{Introduction}

The number of biomedical texts with rich textual information and complex data structures is increasing exponentially. Much important biomedical information and knowledge are hidden in it. 
Medical knowledge graphs (MKGs) represent medical entities and relations in the form of nodes and edges. 
Analyzing these entities, including chemical-disease, chemical–protein, especially the interactions between the pairs,
can be critical to study the potential mechanisms behind massive biological functions.
For example, the electronic health records and pathology reports contain biological information such as biochemical, image, and pathology information in diagnosis and treatment.
Generally, a large-scale MKG contains vast amounts of biomedical entity relations to support healthcare applications. Named Entity Recognition (NER) and Relation Extraction (RE) are promising technologies to automatically build such large-scale MKGs.


The number of new cancer cases and cancer deaths in China ranks first in the world in 2020, Using population-level statistics to monitor cancer prevalence and outcomes is essential in understanding the disease and developing treatment and prevention plans \cite{hayat2007cancer}. 
The pathology report plays a critical role in diagnosing and staging cancer, which usually consists of three parts, the first part is the visual observation of the biopsy tissue, the second part is the professional description of the biopsy tissue at the molecular level, and the third part is the professional diagnosis given by the pathologist. Pathology reports are usually recorded by the pathologists in natural language and stored in a database.
Therefore, the pathologist can analyze the patient's pathological features from the pathology report to guide the prognosis.  
However, due to the flexibility of natural language, the analysis of pathology reports is time-consuming and laborious. Using structured pathology reports can reduce the influence of subjective factors on analysis and effectively improve the efficiency of report analysis.
The design of structured pathology report templates is difficult due to the different standards among different medical organizations. In addition, pathologists prefer to use natural language to record their observations and analysis results freely and flexibly compared to the templates that lack flexibility. Such unstructured data cannot be directly used for data mining and analysis.
Therefore, it is valuable to transform unstructured data into structured data that can be analyzed and processed by computers. Pathologists can combine patients' structured pathology reports with images to perform cluster analysis on patients, which is conducive to the selection and treatment of targeted drugs.

In recent years, the deep neural networks have been widely used in various natural language processing (NLP) tasks showing superior performance, such as named entity recognition, relationship extraction, sentiment classification and machine translation. Some researchers have developed medical information extraction methods based on deep neural networks and achieved good performance, which are mainly based on recurrent neural networks and convolutional neural networks. Lee et al. \cite{lee2018automated} designed a new data model for representing biomarker knowledge from pathology report. Alawad et al. \cite{alawad2020automatic} implemented 2 different multitask learning (MTL) techniques to train a word-level convolutional neural network (CNN) for automatic extraction of cancer data from unstructured text in pathology reports. In our previous work \cite{wu2020structured}, we applied the graph convolutional neural network to extract medical information and generate structured data from unstructured text in pathology reports. However, pathology reports contain lengthy and complicated descriptions, entities across sentences are difficult to detect, and some critical medical information may be lost. Compared with traditional machine learning methods and neural network methods, our experiments have proved that GCN is more effective than other neural network methods, GCN can capture the semantic and syntactic information between remote sentences based on the dependency structure of sentences. However, it is hard to distinguish the relevance of contextual features for the relation extraction in traditional methods. Li et al. \cite{li2017multi} suggested that attention mechanism can capture the most important semantic information for the information extraction task. Vaswani et al. \cite{vaswani2017attention} introduced a multi-head attention mechanism that applied the self-attention mechanism multiple times to capture the relatively important features from different representation subspaces.

To improve cross-sentence relationships extraction and reduce noise data restraint, we explore the Biomedical Information Extraction (BioIE), an information extraction with multi-head attention mechanism enhanced graph convolutional network. The GCN is exploited to encode the dependency structure of the input sentences, the multi-head attention mechanism can automatically assign weight to each edge according to the importance of nodes in the dependency tree. 

We evaluate our model BioIE on two public biomedical relationship extraction tasks (CDR and CPI) and a cross-hospital pan-cancer pathology report corpus, which contain the data from The Cancer Genome Atlas (TCGA) \cite{tomczak2015cancer} and the data from hospital. In the pathological information extraction task, we extract seven key characteristics from pathology reports--cancer type, lateral site, tumor size, histological type, histological grade, TNM stage and lymphatic metastasis. Each cancer characteristic constitutes a different learning task. The result shows that BioIE achieves better clinical performance than traditional methods and deep learning approaches. Furthermore, we evaluate the applicability of BioIE under a transfer learning setting and show that our method achieves promising performance in processing pathology reports from different formats and writing styles.
Our main contributions are as follows:

1.We construct a cross-institutional data set of cancer pathology reports from the TCGA and local hospital.

2. We apply the graph convolutional network to extract biomedical relations which includes the semantic, syntactic, and sequential representation of the sentences.

3. We use the multi-head attention mechanism to reduce noise data and retain valuable long-range information effectively.

The organizational structure of this paper is as follows: The second part includes related work. The third part describes the data set and our proposed model with the experimental process in detail. In the fourth part, we introduce the statistical data and model performance of our data set, Finally we discuss the experimental results and limitations.

\section{Related Work}

\subsection{Relation extraction based on machine learning}
Relation extraction is an essential subfield of natural language processing. Many approaches have been developed to deal with relation extraction, like bootstrapping, unsupervised relation discovery, and supervised classification \cite{zeng2015distant}. One of the popular methods is supervised machine learning based on feature vectors. The previous representative work \cite{kambhatla2004combining} used the Maximum Entropy model to combine various lexical, syntactic, and semantic features and Nguyen et al. \cite{nguyen2014employing} performed word embedding and clustering on adapting feature-based relation extraction systems. Besides feature-based methods, kernel-based methods are commonly used. Zelenko et al.\cite{zelenko2003kernel} integrated kernels with Support Vector Machine and Voted Perceptron learning algorithms to do person-affiliation and organization-location relations. An unsupervised method for relation extraction was proposed by Hasegawa et al. \cite{hasegawa2004discovering}. In Hasegawa’s work, they clustered pairs of named entities based on the similarity of the context.

\subsection{Relation extraction based on neural network}
In recent years, deep learning methods are widely applied in many NLP tasks. Convolutional neural networks (CNNs) and recurrent neural networks (RNNs) are two popular models. In several relation extraction tasks, the neural networks also showed their effectiveness. For example, Liu et al. \cite{liu2013convolution} proved their CNN on relation extraction was state-of-the-art as their coding method has integrated semantic information into the neural network. Lin et al. \cite{lin2016neural} employed CNNs to embed the semantic for the sentence to do relation extraction on the sentence level. Sahu et al. \cite{sahu2019inter} built a labeled edge GCN model based on document-level graphs to do inter-sentence relation extraction. Other relation extraction tasks, which focused on a specific field, could also be completed with a neural network like multichannel convolutional neural network (MCCNN) used for biomedical area \cite{quan2016multichannel}, maximum entropy model and CNN were both employed on chemical–disease relation extraction task \cite{gu2017chemical,lim2018chemical} used recursive neural network for chemical-gene relation extraction.

\begin{table*}[]
\caption{Different types of relation define in structure reports}
\begin{center}
\begin{tabular}{|l|l|l|}
\hline
\textbf{Type} & \textbf{Description}                                             & \textbf{Example}                                                  \\ \hline
Type          & The type and location of cancer                                  & kidney and adrenal gland                                          \\ \hline
Site          & Laterality or resection site of cancer                           & left, radical nephrectomy                                         \\ \hline
Size          & Maximum diameter of the neoplasm                                 & maximum diameter of the neoplasm is 11 cm                         \\ \hline
Subtype       & Histology subtype in WHO guideline                                                & renal cell carcinoma, conventional (clear and granular cell) type \\ \hline
Grade         & Histology grade in WHO guideline                                                 & Fuhrman's nuclear grade varies from grade II to grade IV          \\ \hline
TNM           & Pathological TNM classification and stage                        & TNM Stage: pT3b NX MX                                             \\ \hline
Metas         & lymphatic metastasis is the most common metastasis mode & Regional Lymph Nodes: Negative 0/2                                \\ \hline
\end{tabular}
\end{center}
\end{table*}

\begin{table*}[]
\caption{The cancer pathology report information}
\begin{center}
\begin{tabular}{|l|l|l|l|l|l|l|l|l|l|l|l|l|}
\hline
\textbf{data set}                  & \multicolumn{7}{c|}{\textbf{TCGA}}                          & \multicolumn{5}{c|}{\textbf{TFAH}}          \\ \hline
\textbf{Cancer}                   & Breast & Lung & Kidney & Colon & Prostate & Gastric & Total & Breast & Kidney & Colon & Gastric & Total \\ \hline
\textbf{Type}                     & 1044   & 978  & 894    & 599   & 480      & 443     & 4438  & 680    & 174    & 261   & 283     & 1398  \\ \hline
\textbf{Site}                     & 783    & 848  & 937    & 405   & 472      & 419     & 3864  & 418    & 174    & 249   & 267     & 1108  \\ \hline
\textbf{Subtype}                  & 1070   & 1026 & 941    & 593   & 499      & 445     & 4574  & 283    & 174    & 162   & 165     & 784   \\ \hline
\textbf{Grade}                    & 1081   & 1020 & 730    & 569   & 441      & 435     & 4276  & 394    & 73     & 189   & 229     & 885   \\ \hline
\textbf{Size}                     & 926    & 855  & 812    & 441   & 369      & 477     & 3880  & 568    & 174    & 186   & 192     & 1120  \\ \hline
\textbf{TNM}                      & 1045   & 920  & 836    & 534   & 477      & 415     & 4227  & /      & /      & /     & /       & /     \\ \hline
\textbf{Metas}                    & 674    & 845  & 453    & 379   & 261      & 334     & 2946  & 298    & 24     & 178   & 176     & 676   \\ \hline
\textbf{\# of Reports}            & 1098   & 1027 & 943    & 599   & 500      & 449     & 4616  & 680    & 174    & 267   & 283     & 1404  \\ \hline
\textbf{\# of sentences}          & 9382   & 8743 & 8087   & 5093  & 4550     & 4146    & 40001 & 5720   & 1157   & 2203  & 2447    & 11527 \\ \hline
\textbf{\# of annotate sentences} & 6623   & 6492 & 5803   & 3520  & 2999     & 2968    & 28405 & 2641   & 793    & 1225  & 1312    & 5971  \\ \hline
\end{tabular}

\end{center}
\end{table*}

\subsection{Graph convolution network}
Graph convolution network (GCN) was first introduced by Kipf \& Welling in 2017 \cite{kipf2016semi}, aiming at performing semi-supervised learning semi-supervised learning on graph-struct data. GCN can be used in a wide range of domains like computer vision, NLP, physics, chemistry, and many other domains \cite{zhang2019graph}. To be specific, GCN has been applied to many tasks in the NLP field, like text classification.
Yao et al. \cite{yao2019graph} proposed TextGCN as a model to process the corpus into a graph and learn the word and document embedding. Tensor Graph Convolutional Networks \cite{liu2020tensor} is another method used for text classification to harmonize and integrate heterogeneous information from the graph. Based on textGCN, a heterogeneous graph convolutional network (HeteGCN) was introduced by Ragesh et al. \cite{ragesh2021hetegcn}. 
Li et al. \cite{li2020graph} proposed a graph neural network for diagnosis prediction in the medical field.
Other applications like the type inference on entities and relations \cite{sun2019joint}, inter-sentence relation extraction \cite{sahu2019inter}, and others all prove that GCN is useful when facing the tasks with text.

\subsection{The attention mechanism}
Attention mechanism was first introduced into the field of natural language processing by Bahdanau et al. \cite{bahdanau2014neural}, which achieved great success in machine translation tasks. After that, there are several forms of attention mechanism being developed. For instance, hard attention \cite{xu2015show}, local and global attention \cite{luong2015effective} and multi-head attention \cite{vaswani2017attention}. In recent years, attention mechanism is widely applied to some tasks in the subfield of natural language processing like and relation extraction. Wang et al. \cite{wang2016relation} proposed a convolutional neural network for relation classification depending on two levels of attention. Shen et al. \cite{shen2016attention} explored the word level attention mechanism to improve the discovery of better patterns in heterogeneous contexts.



\section{Data and Method}
\subsection{Data sets}
Apart from the hardware performance and the selection of model, the success of deep learning training largely depends on the number and representativeness of adequate data in the training data set and the consistency of labeling them. Thus, high quality data sets would be significant for model training and testing.
Chemical-disease relations, chemical-protein interactions, and \\protein-protein interactions are three of the most fundamental relation types in complicated MKGs, which significantly benefit various biomedical tasks, such as protein prediction and knowledge graph construction.
We choose two biomedical data set include CDR corpus \cite{li2016biocreative} and Chemprot corpus \cite{kringelum2016chemprot} for training, testing and evaluation. At the same time, we construct a cross-institutional data set of cancer pathology reports. Our construction process is as follows.

\subsubsection{Corpus preparation}
In this study, we mainly focus on six cancers (lung, breast, gastric, colorectal, kidney, and prostate) that are common among patients and have a high mortality rate according to WHO in 2020. We construct a cross-institutional data set of cancer pathology reports from The Cancer Genome Atlas (TCGA) and the department of pathology in the First Affiliated Hospital of Xi'an Jiaotong University (TFAH) and draft the annotation guidelines.
The TFAH data set consists of the unstructured text from 5878 pathology reports covering cancer cases diagnosed in the hospital from 2015 to 2018. 
Since the reports were written in Chinese by the pathologists,it may be difficult for non-Chinese people to understand and affect the applicability of the transfer learning model. Thus, we translated all the reports into English with Google Translation.
In the TFAH, each pathology report identifies the patient ID (we will delete it in the study to ensure patient privacy). 2353 reports related to metastatic tumors were excluded from the study. In the remaining corpus, 2121 duplicate reports with multiple diagnoses were excluded, and the pathological reports with the same case ID were integrated into one report. The final data set consisted of 1404 cancer pathology reports, each corresponding to a unique primary cancer.
After removing invalid reports (including deleting multiple reports of the same patient, deleting reports challenging to identify, and reports of poor quality), we select 4616 pathology reports from the TCGA database, and 1404 pathology reports from the Local Hospital for annotation.

\subsubsection{Corpus annotation}
To increase the quality of our data sets, we would start with a high-quality pathology report corpus.
By annotating pathology reports from different medical institutions, pathologists can ensure the consistency and high quality of the annotation results, requiring detailed annotation guidelines and multiple iterations.

Based on two biomedical corpus annotation examples \cite{gurulingappa2012development,roberts2009building}, the corresponding author and pathologists drafted the first edition of the annotation guideline. We design an iterative annotation workflow by revising our annotation guideline several times. 
To better understand and unify the pathology reports, seven types of cancer variables are defined for extraction, according to the World Health Organization and American Joint Committee on Cancer standard. Table 1 summarizes the variables defined for cancer, including cancer type (Type), tumor resection site (Site), maximum tumor diameter (Size), histology subtype (Subtype), histology grade (Grade), pathological TNM classifications (TNM), lymph node metastasis (Metas). 

The meanings of the variables are introduced as follow: Cancer type refers to the organ where the primary cancer is found. Laterality refers to the side of the corresponding organ on which primary cancer occurs. Size refers to the maximum diameter distance of the tumor. Tumor subtypes describe the types of cells found in cancer tissue. Histological grading is used to determine the rate of cell growth and proliferation. The TNM stage can be used to measure disease progression, which is also instructive to the evaluation and determination of  prognosis. Lymphatic metastasis refers to the movement of cells through lymphatic vessels around the tumor into nearby or even further lymph nodes.
In our annotating session, firstly, Three pathologists annotated the pathology reports in two rounds. We evaluate the annotation consistency with inter-annotator agreement (IAA) metrics and improve the annotation guideline. After completing the training process, six qualified annotators finish annotating the rest pathology reports with the assistance of a rule-based quality control program written by us. All annotations are done at the document-level so that the annotators can leverage the context under challenging cases. An open-source software called MAE version 2.2.10 \cite{rim2016mae2} is used as the annotation tool throughout the entire process.

\subsection{Proposed method}
The structure of our model is shown in figure 1, which includes the following parts: the initialization layer, the Bi-LSTM layer, the multi-head attention layer, the GCN layer, and the relation classification layer. The input of our model is the text sequence, and we first feed the sentence into the initialization layer, which will generate the word representation. The Bi-LSTM layer will obtain contextual features and capture long-range context information from the word representation. The multi-head attention layer applies 
the self-attention mechanism to capture the weighted connections between words in several dependency trees. The GCN layer will guide a contextual representation between word nodes over the document-level dependency graph to capture long-range features.
The final representations of the concatenated vectors are used for relation classification. The details of our model are described in the following sections.

\begin{figure}[!t]
\centering
\includegraphics[width=85mm]{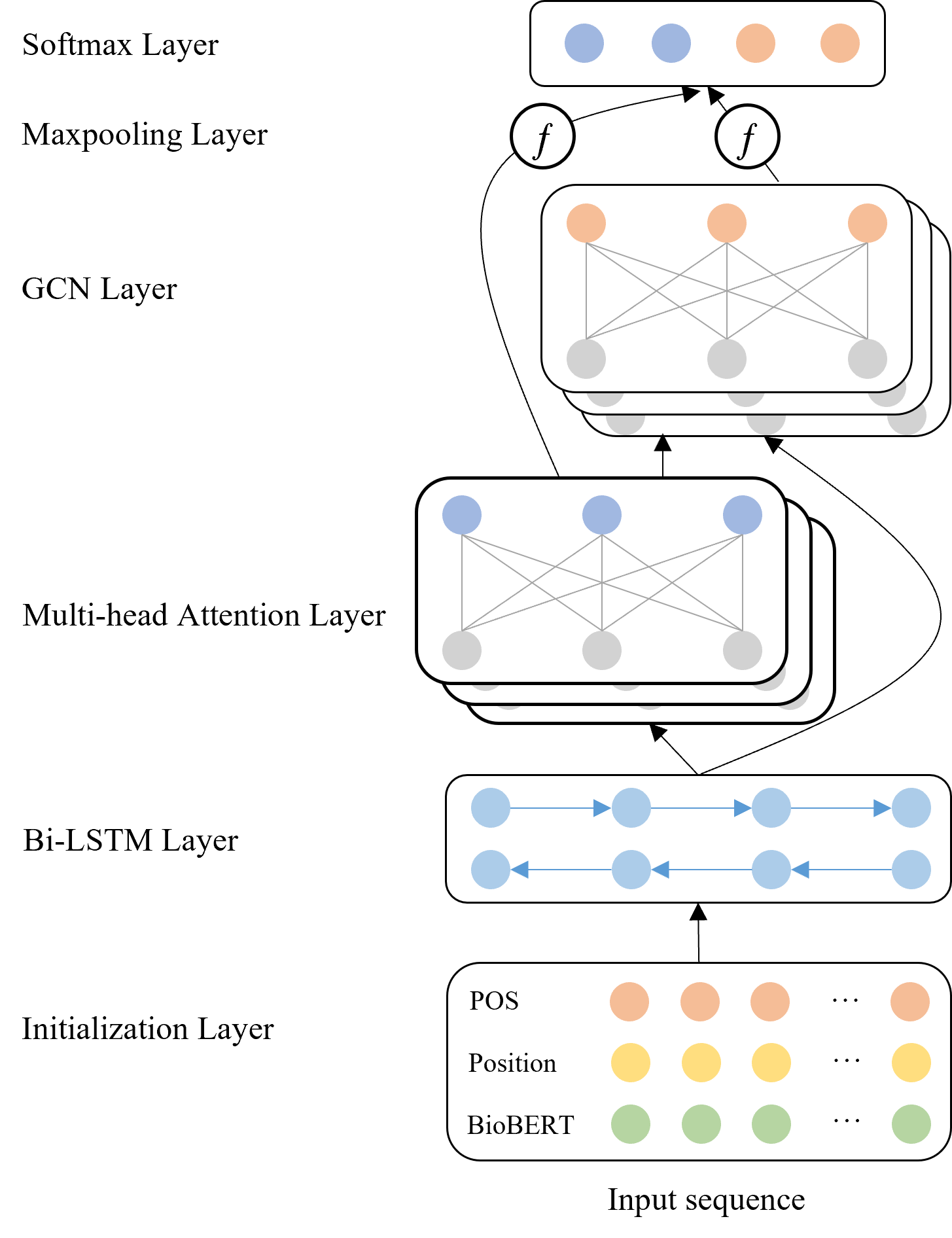}
\caption{The Architecture of Our Work.}
\label{f1}
\end{figure}

\subsubsection{Initialization layer}

Given the sentence $S=\left ( w_{1}, w_{2}, \cdots , w_{n} \right)$, where $n$ is the number of words.
We used the BioBERT word representation \cite{lee2020biobert}, a pre-trained biomedical language representation model, to map our words to the corresponding feature. Also, the position information of tokens are crucial with the relation extraction task. The output of the initialization layer concatenated the vectors of word embedding and position embedding. The final representations are $\omega_{i}=\left [ {w_{i}}^{emb},{w_{i}}^{pos} \right ]$.

\subsubsection{Bi-LSTM layer}

The Bi-LSTM model can better capture the bidirectional semantic dependencies and the long distance dependencies. For the giving sentence 
$S=\left ( \omega_{1}, \omega_{2}, \cdots , \omega_{n} \right)$
,
we represent the Bi-LSTM forward process as follows:
\begin{equation}
\overrightarrow{h_{t}}=LSTM \left (\overrightarrow{h_{t-1}}, \omega_{i} \right)
\end{equation}
\begin{equation}
\overleftarrow{h_{t}}=LSTM \left (\overleftarrow{h_{t-1}}, \omega_{i} \right)
\end{equation}
\begin{equation}
h_{t}= \left[\overrightarrow{h_{t}}\oplus \overleftarrow{h_{t}}\right]
\end{equation}
$ h_{t-1}$ indicates the output of the previous hidden state and $ \omega_{t}$ indicates the input of the current state at the moment $t$. 
$ h_{t}$ is the hidden state of the current time step $t$. $ \overrightarrow{h_{t}}$ and $ \overleftarrow{h_{t}}$ represent the output of the forward LSTM and the backward LSTM, respectively. $\oplus$ is concatenation operation, and the final hidden state $h_{t}$ is the concatenation of forward LSTM and the backward LSTM. The Bi-LSTMs layer can learn the latent features from the input sequences automatically and effectively.

\begin{table*}[]
\caption{Performance comparison with other methods on CDR and CPI corpus}
\begin{center}
\setlength{\tabcolsep}{3.5mm}
\begin{tabular}{|l|l|l|l|l|l|l|l|}
\hline
\multicolumn{4}{|c|}{\textbf{CDR}}                                 & \multicolumn{4}{c|}{\textbf{CPI}}                                  \\ \hline
Methods         & \textbf{P(\%)} & \textbf{R(\%)} & \textbf{F(\%)} & Methods         & \textbf{P(\%)} & \textbf{R(\%)} & \textbf{F(\%)} \\ \hline
Sahu et al. \cite{sahu2019inter}    & 52.8           & 66.0           & 58.6           & Liu et al. \cite{liu2018extracting}    & 57.4             & 48.7           & 52.7           \\ \hline
Lowe et al. \cite{lowe2016efficient}    & 59.3           & 62.3           & 60.8           & Lung et al. \cite{lung2019extracting}     & 63.5           & 51.2           & 56.7           \\ \hline
Zhou et al. \cite{zhou2016exploiting}    & 55.6           & 68.4           & 61.3           & Corbett et al. \cite{corbett2018improving}  & 56.1           & 67.8           & 61.5           \\ \hline
Gu et al. \cite{gu2017chemical}      & 55.7           & 68.1           & 61.3           & Peng et al. \cite{peng2018extracting}     & 72.7           & 57.4           & 64.1           \\ \hline
Proposed method & 61.5           & 72.3           & \textbf{66.4}  & Proposed method & 70.3           & 62.5           & \textbf{66.1}  \\ \hline
\end{tabular}
\end{center}
\end{table*}

\begin{table*}[]
\caption{Detail Indicators For Each Cancer Task}
\begin{center}
\setlength{\tabcolsep}{4mm}
\begin{tabular}{|l|c|c|c|c|c|c|c|c|l|}
\hline
\textbf{Method}           & \multicolumn{3}{c|}{David et al. \cite{martinez2011information} }                                                 & \multicolumn{3}{c|}{John et al. \cite{qiu2017deep}}                                                 & \multicolumn{3}{c|}{Mohammed et al. \cite{alawad2020automatic}}                                               \\ \hline
                 & P(\%)                     & R(\%)                     & F(\%)                     & P(\%)                     & R(\%)                     & F(\%)                     & P(\%)                     & R(\%)                     & \multicolumn{1}{c|}{F(\%)} \\ \hline
\textbf{Cancer}  & 77.5                      & 68.6                      & 72.8                      & 80.6                      & 73.7                      & 76.9                      & 82.0                      & 78.5                      & \multicolumn{1}{c|}{80.2}  \\ \hline
\textbf{Subtype} & 64.9                      & 57.3                      & 60.9                      & 73.2                      & 71.5                      & 72.3                      & 76.8                      & 75.2                      & \multicolumn{1}{c|}{75.9}  \\ \hline
\textbf{Site}    & 72.6                      & 63.6                      & 67.8                      & 75.4                      & 69.3                      & 72.2                      & 76.1                      & 73.6                      & \multicolumn{1}{c|}{74.9}  \\ \hline
\textbf{Size}    & 71.3                      & 67.1                      & 69.1                      & 76.5                      & 74.8                      & 75.6                      & 77.2                      & 75.6                      & \multicolumn{1}{c|}{76.4}  \\ \hline
\textbf{Grade}   & 72.9                      & 66.3                      & 69.4                      & 81.7                      & 80.1                      & 80.9                      & 82.9                      & 80.8                      & \multicolumn{1}{c|}{81.8}  \\ \hline
\textbf{TNM}     & 76.2                      & 72.8                      & 74.5                      & 82.4                      & 81.6                      & 81.9                      & 86.3                      & 85.1                      & \multicolumn{1}{c|}{85.7}  \\ \hline
\textbf{Metas}   & 78.8                      & 73.1                      & 75.8                      & 80.4                      & 78.5                      & 79.4                      & 84.5                      & 83.1                      & 83.8                       \\ \hline
Method           & \multicolumn{3}{c|}{Liu et al. \cite{liu2019bidirectional}}                                             & \multicolumn{3}{c|}{Wu et al. \cite{wu2020structured}}                                                 & \multicolumn{3}{c|}{\textbf{Proposed}}                                             \\ \hline
                 & P(\%)                     & R(\%)                     & F(\%)                     & P(\%)                     & R(\%)                     & F(\%)                     & P(\%)                     & R(\%)                     & \multicolumn{1}{c|}{F(\%)} \\ \hline
\textbf{Cancer}  & \multicolumn{1}{l|}{78.9} & \multicolumn{1}{l|}{74.7} & \multicolumn{1}{l|}{76.7} & \multicolumn{1}{l|}{82.7} & \multicolumn{1}{l|}{80.4} & \multicolumn{1}{l|}{81.5} & \multicolumn{1}{l|}{84.1} & \multicolumn{1}{l|}{82.2} & 83.1                       \\ \hline
\textbf{Subtype} & \multicolumn{1}{l|}{73.1} & \multicolumn{1}{l|}{70.1} & \multicolumn{1}{l|}{71.6} & \multicolumn{1}{l|}{77.4} & \multicolumn{1}{l|}{74.8} & \multicolumn{1}{l|}{76.1} & \multicolumn{1}{l|}{79.3} & \multicolumn{1}{l|}{76.6} & 77.9                       \\ \hline
\textbf{Site}    & \multicolumn{1}{l|}{73.0} & \multicolumn{1}{l|}{69.9} & \multicolumn{1}{l|}{71.4} & \multicolumn{1}{l|}{75.1} & \multicolumn{1}{l|}{73.8} & \multicolumn{1}{l|}{74.4} & \multicolumn{1}{l|}{75.5} & \multicolumn{1}{l|}{73.2} & 74.3                       \\ \hline
\textbf{Size}    & \multicolumn{1}{l|}{73.8} & \multicolumn{1}{l|}{70.1} & \multicolumn{1}{l|}{71.9} & \multicolumn{1}{l|}{78.7} & \multicolumn{1}{l|}{77.3} & \multicolumn{1}{l|}{77.9} & \multicolumn{1}{l|}{77.8} & \multicolumn{1}{l|}{78.4} & 78.1                       \\ \hline
\textbf{Grade}   & \multicolumn{1}{l|}{79.2} & \multicolumn{1}{l|}{76.5} & \multicolumn{1}{l|}{77.8} & \multicolumn{1}{l|}{85.1} & \multicolumn{1}{l|}{84.2} & \multicolumn{1}{l|}{84.6} & \multicolumn{1}{l|}{87.3} & \multicolumn{1}{l|}{85.6} & 86.4                       \\ \hline
\textbf{TNM}     & \multicolumn{1}{l|}{82.9} & \multicolumn{1}{l|}{78.1} & \multicolumn{1}{l|}{80.4} & \multicolumn{1}{l|}{87.6} & \multicolumn{1}{l|}{86.0} & \multicolumn{1}{l|}{86.8} & \multicolumn{1}{l|}{89.4} & \multicolumn{1}{l|}{89.1} & 89.2                       \\ \hline
\textbf{Metas}   & \multicolumn{1}{l|}{86.7} & \multicolumn{1}{l|}{82.3} & \multicolumn{1}{l|}{84.4} & \multicolumn{1}{l|}{88.4} & \multicolumn{1}{l|}{85.7} & \multicolumn{1}{l|}{87.1} & \multicolumn{1}{l|}{88.6} & \multicolumn{1}{l|}{87.8} & 88.2                       \\ \hline
\end{tabular}
\end{center}
\end{table*}

\subsubsection{Multi-head attention layer}

Through the Bi-LSTM layer, our model obtains the context characteristics of the input text. Different features have various weights in biomedical relation extraction tasks. To highlight the relatively important features, we introduce a multi-head attention mechanism so that it can generate different subspaces and reduce the impact of noise data. The essence of multi-head attention is the multiple application of the self-attention mechanism, with which the model learns relatively important features from different representation subspaces.
The hidden output of Bi-LSTM layer is used as the input of multi-head attention layer. Given each input as query $Q$, key $K$, and value $V$, attention scores are obtained by the following scaled dot-product attention calculation method:
\begin{equation}
attention \left( Q,K,V \right)=softmax \left(\frac{QK^{t}}{\sqrt{d}} \right) V
\end{equation}

where $d$ denote the dimension of the output of the hidden units. Therefore, the model can learn more different types of information through the multi-head attention mechanism. 
Considering the multi-head attention contains h heads, the final multi-head attention is the concatenation of each head as $Multihead \left( Q,K,V\right)=Concat \left( head_{1}, head_{2}, \cdots, head_{h}\right)W$.

\subsubsection{GCN layer}

In order to preserve the multi-dimensional information of the text, we use GCN to construct the dependency relation between the words encoded by dependency graphs.
Inspired by the previous work \cite{liu2020tensor}, we constructed three kinds of graph edges on the corpus from three perspectives: semantic-based graph, syntactic-based graph and sequence-based graph. 

\begin{itemize}
\item Semantic-based graph: For each document, we obtain the semantic features of each word from the trained LSTM and calculate the cosine similarity between the words. If the similarity value exceeds the predefined threshold, it means that the two words have a semantic relationship in the current document. We count the number of times each pair of words has a semantic relationship in the entire corpus.
\item Syntax-based graph: For each document in the corpus, we first use a parser to extract the directionless dependencies between words. Similar to the strategy used in semantic graphs, we count the number of times each pair of words has syntactic dependence in the entire corpus and calculate the edge weight of each pair of words.
\item Sequence-based graph: Sequence context describes the language attributes between words. The weight of an edge between two words is the point-wise mutual information (PMI) of them. Using PMI achieves better results than word co-occurrence count in our preliminary experiments.
\end{itemize}

For each layer, we perform two kinds of propagation learning, intra-graph propagation and inter-graph propagation. The GCN layer takes the hidden output of the Bi-LSTM layer as the input. In the layer $l$ of one L-layer GCN, the hidden representation of node $i$ can be calculated as:

\begin{equation}
h_{i}=f \left( \sum_{i=1}^{n}A_{ij}W^{(j)}x_{j}+b^{(j)}\right)
\end{equation}

\begin{equation}
{h_{i}}^{(l)} =f \left( \sum_{i=1}^{n}A_{ij}W^{(l)}{h_{j}}^{(l-1)}/d_{i}+b^{(l)}\right)
\end{equation}
where ${h_{j}}^{(l-1)}$ is the output of the previous GCN layer, $d_{i}$ is the degree of node $i$ in the dependency graph, $b_{(j)}$ and $b_{(l)}$ are bias terms and the activation function $f$ is nonlinear function.

\subsubsection{Relation classification layer}

To extract relations by using the output vector representations of the multi-head attention layer and the GCN layer, we contact the two vectors to form the final representation of relation classification. 
We employ max pooling over the outputs of the multihead attention layer and the GCN layer and then concatenate these 2 vectors as the final representation. 
Probability distribution would be calculated with softmax function.

\subsubsection{Transfer learning}

Transfer learning \cite{goodfellow2016deep} aims to use what has been previously learned from the source domain to learn a better model in the target domain.
In this study, transfer learning techniques are used to reduce the stress and duplication in cross-institutional pathology report analysis. We first train our model on the primary data set (TCGA) and then re-trained the model with transfer parameters by fine-tuning the characteristics learned on the target hospital data set (TFAH), the data set used to train the target task. Then, we exchange the primary data set and the target data set, using TFAH as the original data set, and using the well-trained model to train on the TCGA data set. In our experiment, we do not reduce or change any layers but fine-tuned the parameters in different layers.
We verify the accuracy of different models. If these features are easy to generalize and apply to primary and target tasks, transfer learning can be carried out effectively.

In the experiments, we firstly set the range of parameters according to the experience, then tune the parameters on the validation set, and finally select the model with the optimal parameters to evaluate on the test set.

\section{Experiments}
\subsection{Corpus Statistics}
We first evaluate our proposed method in two major biomedical relationship extraction tasks.
We evaluate our model on the CDR corpus which is the benchmark dataset for the CID relation extraction task. CDR corpus consists of 1500 PubMed abstracts. In this study, the gold entity annotations provided by BioCreative V are used to evaluate our model. 
In addition, we compare our model with other existing methods on another public CPI relation extraction task. We use ChemProt corpus provided by BioCreative VI. The dataset includes 10-type relation classes in which five classes (CPR3,4,5,6,9) are used for evaluation purposes, which totally consists of 7600 CPIs. The PubMed abstracts from biomedical literature constitute the original corpus where more than 98\% of the relation entity pairs are contained in a sentence, and we only need to consider the relation extraction within the sentence. Therefore, different from the CID relationship, we can ignore the cross-sentence entity relationship pairs at the document-level in the CPI relationship extraction, so as to conduct experiments at the 
sentence-level.

In addition, we also evaluate our method on our pathology reports corpus. The original corpus includes a total of 7650 reports collected from the TCGA and TFAH data set with detailed label. After removing invalid data (including missing and duplicate data), 6020 reports are annotated by pathologists and well trained annotators. 
Table 2 lists the detailed statistics of our corpus. As one can see, the size of TCGA is much larger than that of TFAH. Some pathology reports include the patient's label in the title but not in the body of the report. In these cases, we directly associate the patient markers in the pathology report title with the text.
Each pathology report describes an average of 120.3 words. We limit the maximum and minimum lengths of pathology report descriptions to 150 and 50, respectively.

It is worth noting that we retain histological grade (Gleason grade in prostate cancer) and TNM stage as separate types in the type Grade in our experiment. This is because, on the one hand, they represent different meanings of pathological grade and have different clinical guidance roles in structured pathological reports. On the other hand, syntax, co-occurrence, and their order are subtly different from one another. Keeping them as separate types may help improve model performance.

\subsection{Evaluation metrics}
In the experiment, we perform 10-fold cross-validation to test the accuracy of the algorithm. The data set is divided into ten parts, and nine parts are used as training data while the other part is used as test data. 10 percent of the annotation data are randomly selected for verification. 

In order to evaluate different models and classifiers, the standard macro-Precision (P), Recall (R) and F-score (F) are used to evaluate the performance of our model against the gold annotations. TP, FN and FP represent true positive, false negative and false positive, respectively. The F score is a harmonic mean of precision and recall rate. 


For all metrics, by calculating 95\% confidence intervals from the test set, we are able to better determine the statistical significance of the difference in performance between the baseline model and our proposed approach. 

In the experiments, we firstly set the range of parameters according to experience, then use grid search to tune the parameters on the development set, and finally select the model with the optimal parameters to evaluate on the test set.

\begin{table}[]
\caption{Performance Comparison With Other Methods}
\begin{center}
\begin{tabular}{|l|l|l|l|}
\hline
Method            & \textbf{P(\%)} & \textbf{R(\%)} & \textbf{F(\%)} \\ \hline
David et al. \cite{martinez2011information}       & 68.3           & 57.5           & 62.4           \\ \hline
John et al. \cite{qiu2017deep}       & 82.3           & 79.8           & 81.0           \\ \hline
Mohammed et al. \cite{alawad2020automatic}   & 84.6           & 81.2           & 82.9           \\ \hline
Liu et al. \cite{liu2019bidirectional}  & 81.0           & 81.8           & 81.4           \\ \hline
Wu et al. \cite{wu2020structured}      & 85.7           & 83.6           & 84.6           \\ \hline
Proposed method  & 86.9           & 83.7           & \textbf{85.3}  \\ \hline
\end{tabular}
\end{center}
\end{table}

\subsection{Performance}
In this experiment, we compare our model on CID relation extraction and CPI extraction separately with other existing methods. The results shown in Table 3 indicate that our model achieves the best performance on both tasks.

\subsubsection{Performance comparisons on CID relation extraction}
We compare our model with several state-of-the-art methods of the CID relation extraction.
In the left part of Table 3, the neural network methods achieved competitive performance in the CID relation extraction task.

Lowe et al. \cite{lowe2016efficient} developed a rule-based system with specific patterns to extract CID relations. However, the hand-crafted rules is laborious and time-consuming.
Compared with the rule-based approach, machine learning methods have shown a promising capability of  relation extraction. 
Sahu et al \cite{sahu2019inter} proposed a graph convolutional neural network model on a document-level graph.
Zhou et al. \cite{zhou2016exploiting} proposed a hybrid method which consisted an LSTM network with SVM for the sentence-level CID relations. Their model achieves improvement with using additional postprocessing heuristic rules.
Gu et al. \cite{gu2017chemical} proposed a convolution neural network (CNN) model based on contextual and dependency information. They also used additional postprocessing heuristic rules to improve performance. 
Compared with the methods above, our model can effectively learn semantic and syntactic information from complex sentences. And the results suggest that our model achieves the highest F-scores.

\subsubsection{Performance comparisons on CPI extraction}

We compare our model with several state-of-the-art methods of the CPI relation extraction.
In the right part of Table 3, the neural network methods achieved competitive performance in the ChemProt corpus.

Liu et al. \cite{liu2018extracting} synthesized the GRU model with attention pooling.
Corbett et al. \cite{corbett2018improving} used a Bi-LSTMs model with pretrained LSTM layers to extract CPIs.
Lung et al. \cite{lung2019extracting} used a multiple model to integrate the semantic and dependency graph features.
Peng et al. \cite{peng2018extracting} applied majority vote or ensemble method to combine the results of SVM, CNN and Bi-LSTM models. 
Compared with other methods, our model is more effective in biomedical relation extraction by capturing long-range  information.

\begin{table}[]
\caption{Performance Comparison in Transfer Learning}
\begin{center}
\begin{tabular}{|l|l|l|l|l|l|l|}
\hline
          & \multicolumn{3}{c|}{TCGA-TFAH} & \multicolumn{3}{c|}{TFAH-TCGA} \\ \hline
Method                & P(\%)    & R(\%)    & F(\%)    & P(\%)    & R(\%)    & F(\%)    \\ \hline
SVM             & 43.3     & 51.4     & 47.1     & 37.6     & 46.5     & 41.6     \\ \hline
CNN             & 52.7     & 49.1     & 50.8     & 46.9     & 42.1     & 44.4     \\ \hline
MT-CNN          & 61.3     & 52.9     & 56.7     & 51.5     & 48.2     & 49.8     \\ \hline
Bi-LSTM         & 58.1     & 48.7     & 52.8     & 46.6     & 43.7     & 45.1     \\ \hline
GCN             & 67.5     & 56.3     & 61.4     & 60.9     & 51.4     & 55.7     \\ \hline
Proposed method & 75.4     & 58.4     & 65.8     & 61.5     & 54.3     & 56.7     \\ \hline
\end{tabular}
\end{center}
\end{table}

\subsubsection{Performance comparisons on pathological information extraction}
In this experiment, we compare our model on our corpus with other existing methods in Table 4 and Table 5. 
David et al. \cite{martinez2011information} used Naïve-Bayes and support vector machine (SVM) as their classifier in the relation extraction tasks. John et al. \cite{qiu2017deep} indicated that CNN may perform well in information extraction of pathology reports. Mohammed et al. \cite{alawad2020automatic} used multi-task convolutional neural network(MT-CNN) to achieve a better classified effect.  Liu et al. \cite{liu2019bidirectional} used bidirectional LSTM(Bi-LSTM) in text classification. Wu et al. \cite{wu2020structured} proposed a new model that constructs a separate graph for each text-level input. 
The model we use extracts seven different cancer characteristics from cancer pathology reports. Each cancer characteristic constitutes a different learning sub-task. 
Compared with these methods, our model can improve cross-sentence relationships extraction and reduce noise data restraint. The results displayed in Table 5 shows that our model achieves the most advanced performance, F-score of 85.3\%, on the whole corpus. In each sub-task in Table 4, our model also achieves the best accuracy.

Overall, our method takes advantage of the deep context representation, graph convolutional network and multi-head attention mechanism to achieve state-of-the-art performance on the relation extraction task.

In this experiment, we further study the effects of transfer learning methods on cross-hospital data sets.
We first use TCGA as the primary data set since its size is larger than that of TFAH. 
We conduct experiments to examine the effect of transfer knowledge learned from TCGA to TFAH and from TFAH to TCGA.
As we can see from the experimental results in Table 6, the generality and re-usability of the proposed model are improved compared with other models. Compared with Table 5, the transfer learning model can find a more significant performance decline on both data sets. In real-world situations, the format and writing style of descriptive pathology reports differ across hospitals.

\section{Discussion}
\subsection{Ablation study}
To study the contribution of each component in the model, we perform ablation studies on the corpus, where we delete different parts of the model at a time. The model network architecture and results are shown in the Table 7.

First, we evaluate the effectiveness of the word representation of our model. We change the input representations from Word2Vec to BioBert with position embedding.
The results show that the BioBert pre-trained model is able to generate more comprehensive word representations based on sentence context, and combining position embedding with word embedding can further improve performance. When we concatenate the pre-trained word representation and position embedding as the input representation, we achieve the best F-measure of 85.3\%.

When we remove the multihead attention layer and the GCN layer, the F-measure drops by 1.2\% and 2.6\%, respectively. We can observe that removing either the multi-head attention layer or the GCN layer reduces the performance of the model. 
The addition of GCN can capture structural information by using the document graph and effectively capture long-range syntactic features using the dependency structure of input sentences. 

We evaluate the effectiveness of the multi-head attention mechanism. We deal with the output of Bi-LSTM by different attention mechanisms--without attention mechanism, scaled dot-product attention in single-head attention mechanism, and multi-head attention mechanism. From the result we can see that using the multi-head attention mechanism can improve the performance of the relation extraction task.
The multi-head attention mechanism can effectively reduce the impact of noise data without losing valuable information in the sentence. It can learn the global dependency information by modeling the implicit dependency relationship between words, which helps the model produce a better graph representation.

\begin{table}[]
\caption{Ablation study}
\begin{center}
\begin{tabular}{|l|c|}
\hline
\textbf{Model}                                 & \multicolumn{1}{l|}{\textbf{F-Score(\%)}} \\ \hline
Proposed Method                                & 85.3                                      \\ \hline
- BioBert                                      & 77.8                                      \\ \hline
- position                                     & 84.5                                      \\ \hline
- position - BioBert                           & 76.3                                      \\ \hline
- Multi-head Attention                         & 84.1                                      \\ \hline
- Multi-head Attention + Single-head attention & 84.6                                      \\ \hline
- GCN                                          & 82.7                                      \\ \hline
\end{tabular}
\end{center}
\end{table}

\subsection{Error analysis}
By comparing the output of the model to the pathologist's standard, we analyze the final results for errors. There are two main types of errors:
In the misclassification results, we find that most of the samples mistake the non-relational type for the relational type, such as the misclassification of the cancer resection site and the type of cancer. There are also some errors like misclassifying the location of the lump within the organ as the site of excision. According to our analysis of the results, it is possible that the attention mechanism gives higher weight to entities that have across-sentences relationships .
In another case, multiple entities appear in a single sentence and some entities have relationships with each other. This is like the case of a mixed diagnosis of several subtypes. We will consider using better pretreatment and post-processing technology to solve the above problems in future work. 

\section{Conclusion}
In this paper, we proposed Biomedical Information Extraction (BioIE), a hybrid neural network to extract relations from biomedical text and unstructured pathology reports. We used GCN to obtain graph representation based on semantics, syntax, and sequences to improve relationship extraction performance. Using the multi-head attention mechanism, we can effectively reduce the impact of noise data and obtain relatively important contextual features without losing valuable information. Combining multi-head attention with GCN can further improve the performance of the model. The results showed that our model achieved the most advanced performance on the two major biomedical relationship extraction corpus and a cross-hospital pan-cancer pathology report corpus. With evaluating the applicability of BioIE under a transfer learning setting, the results showed that BioIE achieves promising performance in processing pathology reports from different formats and writing styles

The limitation of our work is the initialization of the text representation and the graph representation. In future work, we will integrate pathology knowledge, define pathology ontologies, build pathology knowledge bases, which contain a large number of structured data in the form of triples (entities, relationships, entities), and combine them with powerful pre-training language models to further improve the performance of our approach.


\section*{Acknowledgment}
This work has been supported by National Natural Science Foundation of China (61772409); The consulting research project of the Chinese Academy of Engineering (The Online and Offline Mixed Educational Service System for “The Belt and Road” Training in MOOC China); Project of China Knowledge Centre for Engineering Science and Technology; The innovation team from the Ministry of Education (IRT\_17R86); and the Innovative Research Group of the National Natural Science Foundation of China (61721002). The results shown here are in whole or part based upon data generated by the TCGA Research Network: https://www.cancer.gov/tcga.



%





\end{document}